\documentclass[11pt,a4paper]{article}
\usepackage[hyperref]{emnlp2018}
\usepackage{times}
\usepackage{latexsym}

\usepackage{url}

\usepackage{booktabs}
\usepackage{amsbsy}
\usepackage{footmisc}
\usepackage{amsmath,amsfonts,amsthm,amssymb}
\usepackage{multirow}
\usepackage{graphicx}
\usepackage{thmtools}
\usepackage{thm-restate}
\usepackage{cleveref}
\usepackage{caption}
\usepackage{subcaption}
\usepackage{todonotes}

\usepackage[normalem]{ulem}

\aclfinalcopy % Uncomment this line for the final submission
\def\aclpaperid{1562} %  Enter the acl Paper ID here

\usepackage{algorithmicx} 
\usepackage[noend]{algpseudocode}
\usepackage{algorithm}
\usepackage[T1]{fontenc}
\algnewcommand\algorithmicdefinitions{\textbf{Definitions:}}
\algnewcommand\Definitions{\item[\algorithmicdefinitions]}
\algnewcommand\algorithmicindices{\textbf{Indices:}}
\algnewcommand\Indices{\item[\algorithmicindices]}
\newlength{\commentindent}
\algrenewcommand\alglinenumber[1]{\footnotesize #1:}

\newcommand{\eat}[1]{}

\newcommand{\yae}[2]{{#1}{#2}}

\newcommand{\x}{\mathbf{x}}
\newcommand{\U}{\mathbf{U}}

\newcommand{\h}{\mathbf{h}}
\newcommand{\cc}{\mathbf{c}}
\newcommand{\W}{\mathbf{W}}
\newcommand{\vv}{\mathbf{v}}
\newcommand{\w}{\mathbf{w}}
\newcommand{\f}{\mathbf{f}}
\newcommand{\rr}{\mathbf{r}}
\newcommand{\bb}{\mathbf{b}}

\title{Simple Recurrent Units for Highly Parallelizable Recurrence}

\author{
Tao Lei$^1$ $\ \quad$ Yu Zhang$^2$ $\ \quad$ Sida I. Wang$^{1,3}$ $\ \quad$ Hui Dai$^1$ $\ \quad$ Yoav Artzi$^{1,4}$\\[0.1em]
$^1$ASAPP Inc. $\quad$ $^2$Google Brain $\quad$ $^3$Princeton University $\quad$ $^4$Cornell University \\[0.1em]
\begin{tabular}{c@{~~~~~~~~~~~~~~~~~~}c}
{\tt $^1$\{tao, hd\}@asapp.com } & {\tt $^2$ngyuzh@google.com}\\
{\tt $^3$sidaw@cs.princeton.edu} & {\tt $^4$yoav@cs.cornell.edu}
\end{tabular}
}
\date{}

\begin{document}
\maketitle

\begin{abstract}
Common recurrent neural architectures scale poorly due to the intrinsic difficulty in parallelizing their state computations. In this work, we propose the Simple Recurrent Unit (SRU), a light recurrent unit that balances model capacity and scalability. 
SRU is designed to provide expressive recurrence, enable highly parallelized implementation, and comes with careful initialization to facilitate training of deep models.
% SRU is carefully crafted to enable highly parallelized implementation, and careful initialization to facilitate training of deep models.
% SRU is designed to support highly parallelized implementation, and careful initialization to facilitate training of deep models.
%that simplifies the computation and exposes more parallelism. 
% SRU is crafted with careful architecture design, parallelized implementation as well as proper initialization to facilitate learning.
We demonstrate the effectiveness of SRU on multiple NLP tasks. SRU achieves 5--9x speed-up over cuDNN-optimized LSTM on classification and question answering datasets, and delivers stronger results than LSTM and convolutional models. 
We also obtain an average of 0.7 BLEU improvement over the Transformer model~\citep{vaswani2017attention} on translation by incorporating SRU into the architecture.\footnote{Our code is available at \url{https://github.com/taolei87/sru}.}
%without additional cost on parallelization.

%In SRU, the majority of computation for each step is independent of the recurrence and can be easily parallelized. SRU is as fast as a convolutional layer and 5-10x faster than an optimized LSTM implementation.
%We study SRUs on a wide range of applications,  including classification, question answering, language modeling, translation and speech recognition. Our experiments demonstrate the effectiveness of SRU and the trade-off it enables between speed and performance. 
%For instance, the forward pass computation of $h_t$ is blocked until the entire computation of $h_{t-1}$ finishes, which is a major bottleneck for parallel computing.
%The proposed recurrent unit operates as fast as a convolutional layer and 5-10x faster than cuDNN-optimized LSTM.
%We demonstrate the unit's effectiveness across a wide range of applications including classification, question answering, language modeling, translation and speech recognition.
%We open source our implementation in PyTorch and CNTK.%\footnote{\url{https://github.com/taolei87/sru}}.
\end{abstract}

\section{Introduction}
\label{sec:intro}

%%%%%
% re-thinking intro:
% para 1: rnns are all over the place... in this paper
% para 2: the speed problems of rnns have been studied extensively and different architectures proposed, but lstm (slow and with many params) is still the most commonly used
% para 3: we study an architecture that retains some level of recurrence, but optimized for common GPUs. it's fast, and light
% para 4: we experiment across multiple challenging domains. with classification and QA, we study the effect of speed, with MT transformer we study the re-introduction of recurrence to SOTA MT. we also study initilization, and show ... 
% para 5: our results ... our implementation for pytorch will be made available 

%%%%%%%%%%%%%%%%%

Recurrent neural networks (RNN) are at the core of state-of-the-art approaches for a large number of natural language tasks, including machine translation~\citep{cho-al-emnlp14,Bahdanau:14neuralmt,jean15,luong2015:EMNLP}, language modeling~\citep{zaremba2014recurrent,Gal2016Theoretically,zoph2016neural}, opinion mining~\citep{Irsoy:14opinion}, and situated language understanding~\citep{Mei:16neuralnavi,Misra:17instructions,Suhr:18context,Suhr:18situated}.
Key to many of these advancements are architectures of increased capacity and computation. 
For instance, the top-performing models for semantic role labeling and translation use eight recurrent layers, requiring days to train~\citep{he2017deep,wu2016gnmt}.
%For example, state-of-the-art semantic role labeling and translation models use eight recurrent layers, and require days to train~\citep{he2017deep,wu2016gnmt}.
The scalability of these models has become an important problem that impedes NLP research.

%The difficulty to scale recurrent neural networks comes naturally from the state computation. 
%RNNs are hard to scale due to the time  time dependence of state computation.
The difficulty of scaling recurrent networks arises from the time dependence of state computation.
In common architectures, such as Long Short-term Memory~\cite[LSTM;][]{hochreiter1997long} and Gated Recurrent Units~\cite[GRU;][]{cho-al-emnlp14}, the computation of each step is suspended until the complete execution of the previous step. 
This sequential dependency makes recurrent networks significantly slower than other operations, and limits their applicability. 
%As a consequence, the applicability of recurrent neural networks can be limited.
\yae{For example, recent translation models consist of non-recurrent components only, such as attention and convolution, to scale model training~\citep{gehring2017convolutional,vaswani2017attention}.}{}
%Different strategies have been proposed to improve the efficiency of RNNs, including for example hardware cache utilization~\citep{diamos2016persistent} and factorization tricks~\citep{kuchaiev2017factorization}.
%Different architectures have been proposed to substitute (part of) recurrence with other components that expose more parallelism~\citep{BalduzziG16}.
%For instance, Quasi-RNNs~\cite{bradbury2016quasi} alternates between convolutions and a minimal recurrent pooling operation for sequence processing.
%More recently, \citet{gehring2017convolutional} and \citet{vaswani2017attention} have employed completely non-recurrent architectures for machine translation.
%Despite considerable efforts~\citep{diamos2016persistent,cudnnlstm,kuchaiev2017factorization}, recurrent networks remain a less favorable choice for large-scale problems.

\begin{figure*}[!ht]
\centering
\includegraphics[width=5.3in]{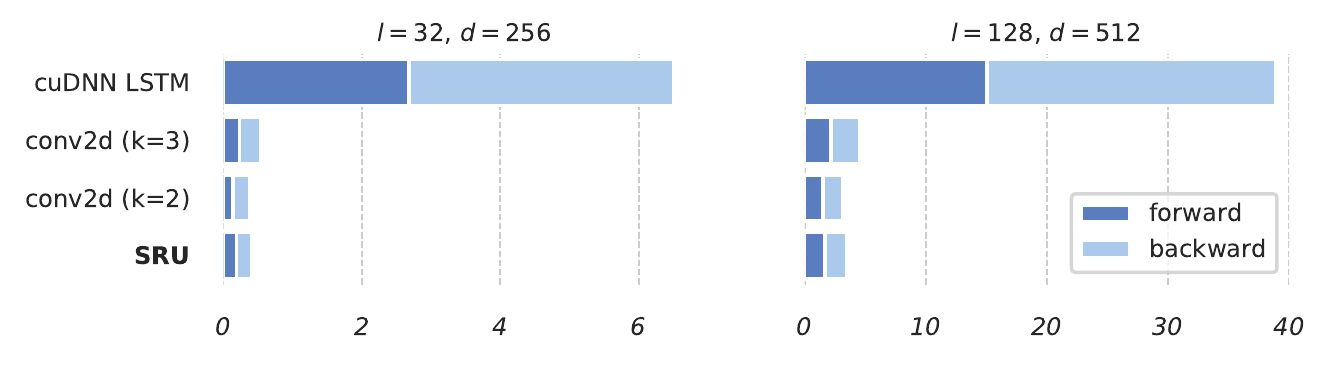}
\vspace{-0.1in}
\caption{Average processing time in milliseconds of a batch of 32 samples using cuDNN LSTM, word-level convolution \texttt{conv2d} (with filter width $k=2$ and $k=3$), and the proposed SRU. We vary the number of tokens per sequence ($l$) and feature dimension ($d$).}
\label{fig:overview}
\end{figure*}

In this work, we introduce the Simple Recurrent Unit (SRU), a unit with light recurrence that offers both high parallelization and sequence modeling capacity.
The design of SRU is inspired by previous efforts, \yae{such as Quasi-RNN~\cite[QRNN;][]{bradbury2016quasi} and Kernel NN~\citep[KNN;][]{lei2017deriving}}{}, but enjoys additional benefits:
\begin{itemize}
\item SRU exhibits the same level of parallelism as convolution and feed-forward nets. This is achieved by balancing sequential dependence and independence: while the state computation of SRU is time-dependent, each state dimension is independent. \yae{This simplification enables CUDA-level optimizations that parallelize the computation across hidden dimensions and time steps, effectively using the full capacity of modern GPUs.}{}
\yae{Figure~\ref{fig:overview} compares our architecture's runtimes to common architectures.}{} 
\item \yae{SRU replaces the use of convolutions (i.e., n-gram filters), as in QRNN and KNN, with more recurrent connections}{}.
This retains modeling capacity, while using less computation (and hyper-parameters).
\item SRU improves the training of deep recurrent models by employing highway connections~\citep{srivastava2015training} \yae{and a parameter initialization scheme tailored for gradient propagation}{} in deep architectures.
%~\citep{wu-zhang-zong:2016:COLING,kim2016character}
%\item SRU employs highway connections, which give rise to improved performance when training deep models~\citep{wu-zhang-zong:2016:COLING,kim2016character}
\end{itemize}
%\yae{}{Besides careful design choices, we make implementation level efforts to improve the training scalability of SRU. We perform GPU / CUDA programming to optimize the computation across dimensions.}
%\yae{}{Figure~\ref{fig:overview} compares our architecture's runtimes to common architectures.} 
%In addition, we derive a parameter initialization specifically tailored for SRU, resulting in improved training performance.

We evaluate SRU on a broad set of problems, including text classification, question answering, translation and character-level language modeling.
Our experiments demonstrate that light recurrence is sufficient for various natural language tasks, offering a good trade-off between scalability and representational power.
On classification and question answering datasets, SRU outperforms common recurrent and non-recurrent architectures, while achieving 5--9x speed-up compared to cuDNN LSTM.
Stacking additional layers further improves performance, while incurring relatively small costs owing to the cheap computation of a single layer. 
We also obtain an average improvement of 0.7 BLEU score on the English to German translation task by incorporating SRU into Transformer~\citep{vaswani2017attention}.

\section{Related Work}
\label{sec:related}

Improving on common architectures for sequence processing has recently received significant attention~\citep{greff2015lstm,BalduzziG16,miao2016simplifying,zoph2016neural,lee2017recurrent}. 
One area of research involves incorporating word-level convolutions (i.e. n-gram filters) into recurrent computation~\citep{lei:2015:EMNLP,bradbury2016quasi,lei2017deriving}.
For example, Quasi-RNN~\citep{bradbury2016quasi} proposes to alternate convolutions and a minimalist recurrent pooling function and achieves significant speed-up over LSTM.
%Significant speed advantages have been demonstrated utilizing parallelized convolution and recurrent pooling computation.
%\citet{lei2017deriving} find that light recurrence can produce competitive results through layer stacking, and study the theoretical aspects of these architectures.
%SRU is built on top of previous efforts, but further optimizes the architecture and computation.
\yae{While \citet{bradbury2016quasi} focus on the speed advantages of the network, \citet{lei2017deriving} study the theoretical characteristics of such computation and possible extensions.}{}
\yae{Their results suggest that simplified recurrence retains strong modeling capacity through layer stacking.}{}
\yae{This finding motivates the design of SRU for both high parallelization and representational power.}{}
SRU also relates to IRNN~\citep{le2015simple}, which uses an identity diagonal matrix to initialize hidden-to-hidden connections.
SRU uses point-wise multiplication for hidden connections, which is equivalent to using a diagonal weight matrix.
This can be seen as a constrained version of diagonal initialization.

Various strategies have been proposed to scale network training~\citep{goyal2017accurate} and to speed up recurrent networks~\citep{diamos2016persistent,shazeer2017outrageously,kuchaiev2017factorization}.
For instance, \citet{diamos2016persistent} utilize hardware infrastructures by stashing RNN parameters on cache (or fast memory).
\citet{shazeer2017outrageously} and \citet{kuchaiev2017factorization} improve the computation via conditional computing and matrix factorization respectively.
Our implementation for SRU is inspired by the cuDNN-optimized LSTM~\citep{cudnnlstm}, but enables more parallelism -- while cuDNN LSTM requires six optimization steps, SRU achieves more significant speed-up via two optimizations.
%The topic of improving learning times was also studied. For example, \citet{goyal2017accurate} addressed stability issues of distributed training with large mini-batches to improve training time. 
%Our approach can be combined with such training procedures. 

The design of recurrent networks, such as SRU and related architectures, raises questions about representational power and interpretability~\citep{chen2017recurrent,peng2018emnlp}.
\citet{BalduzziG16} applies type-preserving transformations to discuss the capacity of various simplified RNN architectures. 
Recent work~\citep{anselmi2015deep,DanielyFS16,zhang16l1,lei2017deriving} relates the capacity of neural networks to deep kernels.
%\citet{lei2017deriving} shows that a broad model class, including SRU and word-level CNN,  can be seen as embedding sequence similarity functions such as string kernels~\citep{lodhi2002text}.
%Layer stacking can then be interpreted as using higher-order sequence similarities, which introduces more non-linearity and representational power. 
We empirically demonstrate SRU can achieve compelling results by stacking multiple layers.

\section{Simple Recurrent Unit}
\label{sec:sru}
%\tl{looks worse than not having the subsecion here?}
%\subsection{Unit architecture}
We present and explain the design of Simple Recurrent Unit (SRU) in this section. 
A single layer of SRU involves the following computation:
\begin{align}
\f_t\ &=\ \sigma\left(\W_f \x_t + \vv_f\odot \cc_{t-1} + \bb_f\right) \\
\cc_t\ &=\ \f_t\odot \cc_{t-1} + (1-\f_t)\odot (\W \x_t) \\[0.3em]
\rr_t\ &=\ \sigma\left(\W_r\x_t + \vv_r\odot \cc_{t-1} + \bb_r\right) \\
\h_t\ &=\ \rr_t\odot \cc_t + (1-\rr_t)\odot \x_t 
\end{align}
where $\W$, $\W_f$ and $\W_r$ are parameter matrices and $\vv_f$, $\vv_r$, $\bb_f$ and $\bb_v$ are parameter vectors to be learnt during training.
%$\h_t$ and $\cc_t$ are the output state and internal state of SRU at step $t$, $\f_t$ and $\rr_t$ are two sigmoid gates, and finally $g()$ denotes an optional activation function.
The complete architecture decomposes to two sub-components: a \emph{light recurrence} (Equation 1 and 2) and a \emph{highway network} (Equation 3 and 4).

The light recurrence component successively reads the input vectors $\x_t$ and computes the sequence of states $\cc_t$ capturing sequential information.
%\begin{align*}
%\cc_t\ &=\ \f_t\odot \cc_{t-1} + (1-\f_t)\odot (\W \x_t) \\
%\f_t\ &=\ \sigma\left(\W_f \x_t + \vv_f\odot \cc_{t-1} + \bb_f\right) .
%\end{align*}
The computation resembles other recurrent networks such as LSTM, GRU and RAN~\citep{lee2017recurrent}.
Specifically, a forget gate $\f_t$ controls the information flow (Equation 1) and the state vector $\cc_t$ is determined by adaptively averaging the previous state $\cc_{t-1}$ and the current observation $\W\x_t$ according to $\f_t$ (Equation 2).
% and alleviates vanishing and exploding gradient problems.

One key design decision that differs from previous gated recurrent architectures is the way $\cc_{t-1}$ is used in the sigmoid gate.
Typically, $\cc_{t-1}$ is multiplied with a parameter matrix to compute $\f_t$, e.g., $\f_t=\sigma(\W_f\x_t + \mathbf{V}_f\cc_{t-1}+\bb_f)$.
However, the inclusion of $\mathbf{V}_f\cc_{t-1}$ makes it difficult to parallelize the state computation: each dimension of $\cc_t$ and $\f_t$ depends on all entries of $\cc_{t-1}$, and the computation has to wait until $\cc_{t-1}$ is fully computed.
To facilitate parallelization, our light recurrence component uses a point-wise multiplication $\vv_f\odot \cc_{t-1}$ instead. 
With this simplification, each dimension of the state vectors becomes independent and hence parallelizable.
%We describe more details of speed optimization in Section 3.x

The highway network component~\cite{srivastava2015training} facilitates gradient-based training of deep networks.
It uses the reset gate $\rr_t$ (Equation 3) to adaptively combine the input $\x_t$ and the state $\cc_t$ produced from the light recurrence (Equation 4),
%\begin{align*}
%\h_t\ &=\ \rr_t\odot \cc_t + (1-\rr_t)\odot \x_t \\
%\rr_t\ &=\ \sigma\left(\W_r\x_t + \vv_r\odot \cc_{t-1} + \bb_r\right) , 
%\end{align*}
where $(1-\rr_t)\odot \x_t$ is a skip connection that allows the gradient to directly propagate to the previous layer.
Such connections have been shown to improve scalability~\cite{wu-zhang-zong:2016:COLING,kim2016character,he2016deep,ZillySKS17}.

%The complete architecture of SRU is:
%\begin{align}
%\cc_t\ &=\ \f_t\odot \cc_{t-1} + (1-\f_t)\odot (\W \x_t) \\
%\h_t\ &=\ \rr_t\odot g(\cc_t) + (1-\rr_t)\odot \x_t \\[0.5em]
%\f_t\ &=\ \sigma\left(\W_f \x_t + \vv_f\odot \cc_{t-1} + \bb_f\right) \\
%\rr_t\ &=\ \sigma\left(\W_r\x_t + \vv_r\odot \cc_{t-1} + \bb_r\right)
%\end{align}
%where $\{\h_t\}$ are the output state vectors that can be passed to subsequent layers.
The combination of the two components makes the overall architecture simple yet expressive, and easy to scale due to enhanced parallelization and gradient propagation.
%Next, we detail our implementation efforts to further optimize the training scalability.

\subsection{Parallelized Implementation}
\label{sec:method:cuda}

Despite the parallelization friendly design of SRU, a naive implementation which computes equations (1)--(4) for each step $t$ sequentially would not achieve SRU's full potential. %achieve significant enough speed improvement. 
%Inspired by cuDNN LSTM implementation~\citep{cudnnlstm}, 
We employ two optimizations to enhance parallelism. 
The optimizations are performed in the context of GPU / CUDA programming, but the general idea can be applied to other parallel programming models.

%Optimizing SRU is similar to how LSTM is optimized in cuDNN~LSTM~\citep{cudnnlstm}. 
%The SRU formulation permits two optimizations.
We re-organize the computation of equations (1)--(4) into two major steps.
First, given the input sequence $\{\x_1\cdots \x_L\}$, we batch the matrix multiplications across all time steps.
This significantly improves the computation intensity (e.g. GPU utilization).
The batched multiplication is:
\begin{align}
\nonumber 	\U^\top &= \left(\begin{array}{l}\W \\ \W_f \\ \W_r \end{array}\right) [\x_1, \x_2, \cdots, \x_L] \;\;,
\end{align}
where $L$ is the sequence length, $\U \in \mathbb{R}^{L\times 3d}$ is the computed matrix and $d$ is the hidden state size.
When the input is a mini-batch of $B$ sequences, $\U$ would be a tensor of size $(L, B, 3d)$. 
%We choose a length-major representation instead of a batch-major version here.%\ya{any specific reason}.

\begin{algorithm*}[!ht!]
\fontsize{10.5}{13.5}\selectfont
%\setstretch{1.}
\caption{\strut Mini-batch version of the forward pass defined in Equations~(1)--(4).}
\label{alg:train}
\begin{algorithmic}
\Indices Sequence length $L$, mini-batch size $B$, hidden state dimension $d$.
\Require Input sequences batch $\x[l,i,j]$; grouped matrix multiplication $\U[l,i,j']$; \\
initial state $\cc_0[i,j]$; parameters $\vv_f[j]$, $\vv_r[j]$, $\bb_f[j]$ and $\bb_r[j]$.
\Ensure Output $\h[\cdot,\cdot,\cdot]$ and internal $\cc[\cdot,\cdot,\cdot]$ states.
\State Initialize $\h[\cdot,\cdot,\cdot]$ and $\cc[\cdot,\cdot,\cdot]$ as two $L\times B\times d$ tensors.
\For{$i=1,\cdots,B; j=1,\cdots, d$}\Comment{\bf $ $ Parallelize each example $i$ and dimension $j$}
	\State $c = \cc_0[i,j]$
	\For{$l=1, \cdots, L$}
    	\State $f = \sigma\left(\,\U[l,i,j+d] + \vv_f[j]\times c + \bb_f[j]\,\right)$ %\Comment{Forget gate}
        \State $c = f\times c + (1-f)\times\U[l,i,j]$ %\Comment{Current internal state}
        \State $r = \sigma\left(\,\U[l,i,j+d\times 2] + \vv_r[j]\times c + \bb_r[j]\,\right)$ %\Comment{Reset gate}
        \State $h = r\times c + (1-r)\times\x[l,i,j]$ %\Comment{Current output state}
        \State $\cc[l,i,j] = c$
        \State $\h[l,i,j] = h$
    \EndFor
\EndFor
\Return $\h[\cdot,\cdot,\cdot]$ and $\cc[\cdot,\cdot,\cdot]$
\end{algorithmic}
\end{algorithm*}

The second step computes the remaining point-wise operations.
Specifically, we compile all point-wise operations into a single fused CUDA kernel and parallelize the computation across each dimension of the hidden state.
Algorithm~\ref{alg:train} shows the pseudo code of the forward function. 
The complexity of this step is $O(L\cdot B\cdot d)$ per layer, where $L$ is the sequence length and $B$ is the batch size.
In contrast, the complexity of LSTM is $O(L\cdot B\cdot d^2)$ because of the hidden-to-hidden multiplications (e.g. $\mathbf{V}\h_{t-1}$), and each dimension can not be independently parallelized.
The fused kernel also reduces overhead. 
Without it, operations such as sigmoid activation would each invoke a separate function call, adding kernel launching latency and more data moving costs.

The implementation of a bidirectional SRU is similar: the matrix multiplications of both directions are batched, and the fused kernel handles and parallelizes both directions at the same time.

\subsection{Initialization}
\label{sec:method:init}

Proper parameter initialization can reduce gradient propagation difficulties and hence have a positive impact on the final performance.
We now describe an initialization strategy tailored for SRU.
%The gated recurrence and highway connections require an adjustment of the common initializations derived for feed-forward networks~\citep{glorot2010understanding,he2015delving}.

We start by adopting common initializations derived for feed-forward networks~\citep{glorot2010understanding,he2015delving}.
The weights of parameter matrices are drawn with zero mean and $1/d$ variance, for instance, via the uniform distribution $[-\sqrt{3/d}, +\sqrt{3/d}]$.
This ensures the output variance remains approximately the same as the input variance after the matrix multiplication.

However, the light recurrence and highway computation would still reduce the variance of hidden representations by a factor of $1/3$ to $1/2$:
\begin{align*}
\frac{1}{3}\ \leq\ \frac{\text{Var}[\h_t]}{\text{Var}[\x_t]}\ \leq\ \frac{1}{2}\;\;,
\end{align*}
and the factor converges to $1/2$ in deeper layers (see Appendix~\ref{sec:appendix:init}).
This implies the output $\h_t$ and the gradient would vanish in deep models.
To offset the problem, we introduce a \emph{scaling correction} constant $\alpha$ in the highway connection
\begin{align*}
\h_t\ &=\ \rr_t\odot \cc_t\, +\, (1-\rr_t)\odot \x_t \cdot \alpha \;\;,
\end{align*}
where $\alpha$ is set to $\sqrt{3}$ such that $\text{Var}[\h_t] \approx \text{Var}[\x_t]$ at initialization. 
When the highway network is initialized with a non-zero bias $\bb_r = b$, the scaling constant $\alpha$ can be accordingly set as:
\begin{align*}
\alpha\ &=\ \sqrt{1+\exp(b)\times 2}\;\;.
\end{align*}
Figure~\ref{fig:init} compares the training progress with and without the scaling correction.
See Appendix~\ref{sec:appendix:init} for the derivation and more discussion.

\begin{figure}[!t!]
\centering
\includegraphics[width=2.8in]{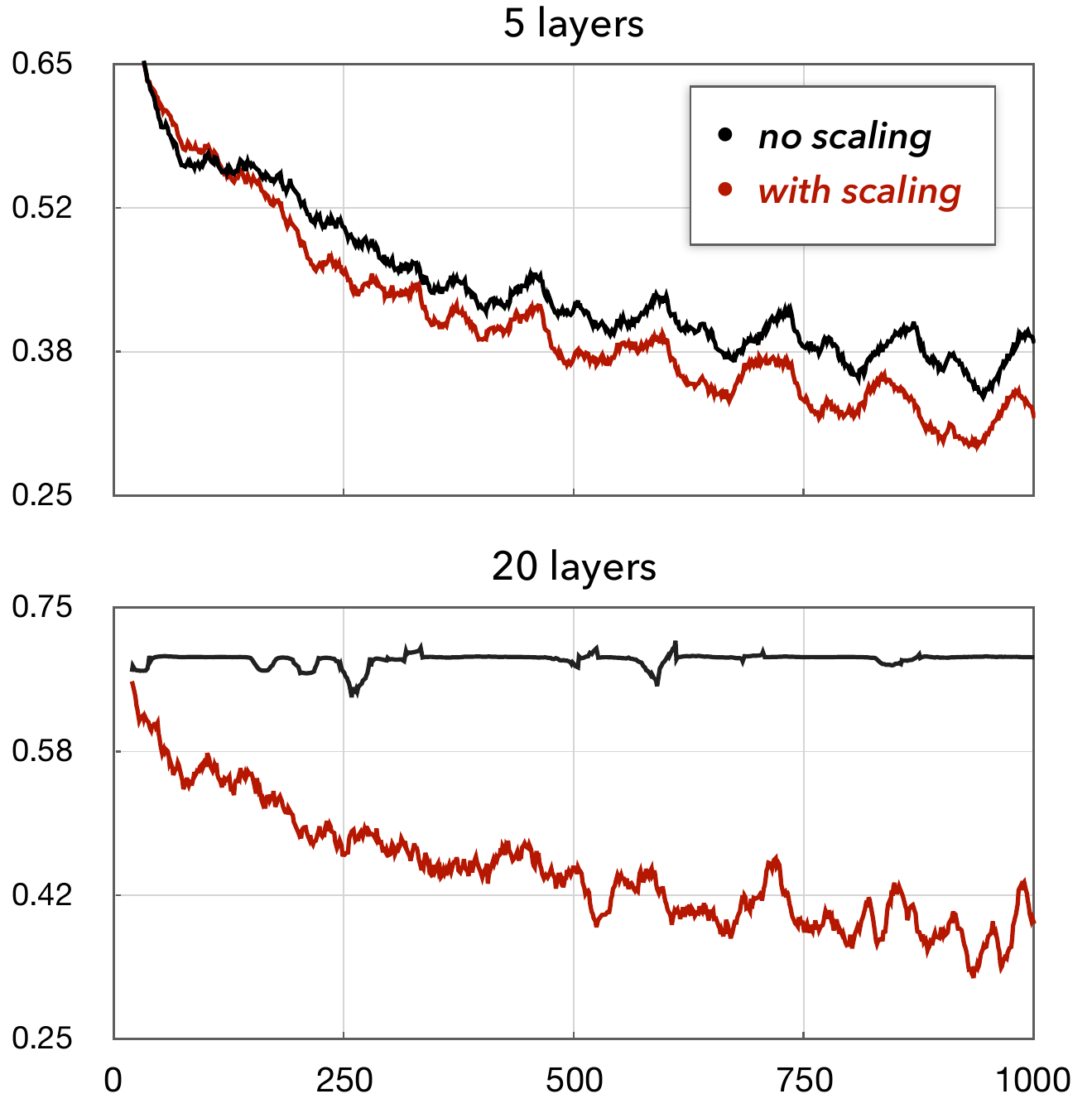}
%\vspace{-0.05in}
\caption{Training curves of SRU on classification. The x-axis is the number of training steps and the y-axis is the training loss. Scaling correction improves the training progress, especially for deeper models with many stacked layers.}
\label{fig:init}
\end{figure}

\begin{table*}[!ht!]
\fontsize{9.8}{12}\selectfont
%\vspace{0.05in}
  \centering
  \begin{tabular}{@{}l@{}c@{~~~~~}ccccccr@{}}
    \toprule
    Model & Size & CR & SUBJ & MR & TREC & MPQA & SST & Time\\
    \midrule
    \multicolumn{7}{@{}l}{\bf Best reported results:} \\[0.3em]
%    \midrule
    \citet{wang2013fast} & & 82.1 & 93.6 & 79.1 & - & 86.3 & - & -\\
    \citet{kalchbrenner:2014} & & - & - & - & 93.0 & - & 86.8 & -\\
    \citet{Kim14} & & 85.0 & 93.4 & 81.5 & 93.6 & 89.6 & 88.1  & -\\
    \citet{zhang2015sensitivity} & & 84.7 & 93.7 & 81.7 & 91.6 & 89.6 & 85.5 & - \\
    \citet{zhao2015self} & & 86.3 & \textbf{95.5} & \textbf{83.1} & 92.4 & \textbf{93.3} & - & - \\
    \midrule
    \multicolumn{7}{@{}l}{\bf Our setup (default Adam, fixed word embeddings):}\\[0.4em]
    CNN & 360k & 83.1$\pm$1.6 & 92.7$\pm$0.9 & 78.9$\pm$1.3 & 93.2$\pm$0.8 & 89.2$\pm$0.8& 85.1$\pm$0.6 & 417\\
    LSTM & 352k & 82.7$\pm$1.9 & 92.6$\pm$0.8 & 79.8$\pm$1.3 & 93.4$\pm$0.9& 89.4$\pm$0.7& 88.1$\pm$0.8 & 2409\\
    QRNN (k=1) & 165k & 83.5$\pm$1.9 & 93.4$\pm$0.6 & 82.0$\pm$1.0 & 92.5$\pm$0.5 & 90.2$\pm$0.7 & 88.2$\pm$0.4 & 345\\
%    QRNN (k=2) & 330k & 84.0$\pm$1.8 & 93.2$\pm$0.8 & 81.6$\pm$1.0 & 93.7$\pm$1.0 & 89.8$\pm$0.8 &  89.5$\pm$0.5 & 381\\[0.5em]
    QRNN (k=1) + highway & 204k & 84.0$\pm$1.9 & 93.4$\pm$0.8 & 82.1$\pm$1.2 & 93.2$\pm$0.6 & 89.6$\pm$1.2 & 88.9$\pm$0.2 & 371\\[0.5em]
   	\textbf{SRU} (2 layers) & 204k & 84.9$\pm$1.6 & 93.5$\pm$0.6 & 82.3$\pm$1.2 & 94.0$\pm$0.5& 90.1$\pm$0.7 & 89.2$\pm$0.3 & \textbf{320}\\
    \textbf{SRU} (4 layers) & 303k & 85.9$\pm$1.5 & 93.8$\pm$0.6 & 82.9$\pm$1.0 & \textbf{94.8}$\pm$0.5& 90.1$\pm$0.6 & \textbf{89.6}$\pm$0.5 & 510\\
    \textbf{SRU} (8 layers) & 502k & \textbf{86.4}$\pm$1.7 & 93.7$\pm$0.6 & \textbf{83.1}$\pm$1.0 & 94.7$\pm$0.5& 90.2$\pm$0.8 & 88.9$\pm$0.6 & 879\\
    \bottomrule
   \end{tabular}
   %\vspace{0.02in}
   \caption{\label{table:clf} Test accuracies on classification benchmarks (Section~\ref{sec:exp:class}).
The first block presents best reported results of various methods. 
The second block compares SRU and other baselines given the same setup.
   For the SST dataset, we report average results of 5 runs.
   For other datasets, we perform 3 independent trials of 10-fold cross validation (3$\times$10 runs).
   The last column compares the wall clock time (in seconds) to finish 100 epochs on the SST dataset.
}
\end{table*}

\section{Experiments}
\label{sec:exp}

We evaluate SRU on several natural language processing tasks and perform additional analyses of the model.
The set of tasks includes text classification, question answering, machine translation, and character-level language modeling. 
Training time on these benchmarks ranges from  minutes (classification) to  days (translation), providing a variety of computation challenges.
%The experiments are performed on PyTorch 0.3\footnote{We found v0.3 has a noticeable speed improvement over previous versions.}. 
%and we set $g(x)=x$ (i.e. identity activation) across our experiments, unless specified otherwise.

%\ya{overview of experiments and pointers to main results and ablations (e.g., classification and QA are direct comparison to LSTM, MT is about adding recurrence without cost to parallelism}

The main question we study is the performance-speed trade-off SRU provides in comparison to other architectures. 
We stack multiple layers of SRU to directly substitute other recurrent, convolutional or feed-forward modules. We minimize hyper-parameter tuning and architecture engineering for a fair comparison. Such efforts have a non-trivial impact on the results, which are beyond the scope of our experiments. 
Unless noted otherwise, the hyperparameters are set identical to prior work.

\subsection{Text Classification}
\label{sec:exp:class}

\paragraph{Dataset} We use six sentence classification benchmarks: %\footnote{\url{https://github.com/harvardnlp/sent-conv-torch}}
movie review sentiment~\citep[MR;][]{pang2005seeing}, sentence subjectivity ~\citep[SUBJ;][]{pang2004sentimental}, customer reviews polarity~\citep[CR;][]{hu2004mining}, question type~\citep[TREC;][]{li2002learning}, opinion polarity~\citep[MPQA;][]{wiebe2005annotating}, and the Stanford sentiment treebank~\citep[SST;][]{socher2013sentiment}.\footnote{We use the binary version of SST dataset.}
%All datasets contain several thousand examples.

Following~\citet{Kim14}, we use word2vec embeddings trained on $100$ billion Google News tokens. 
For simplicity, all word vectors are normalized to unit vectors and are fixed during training.

\begin{figure*}[!th!]
\centering
\includegraphics[width=5.9in]{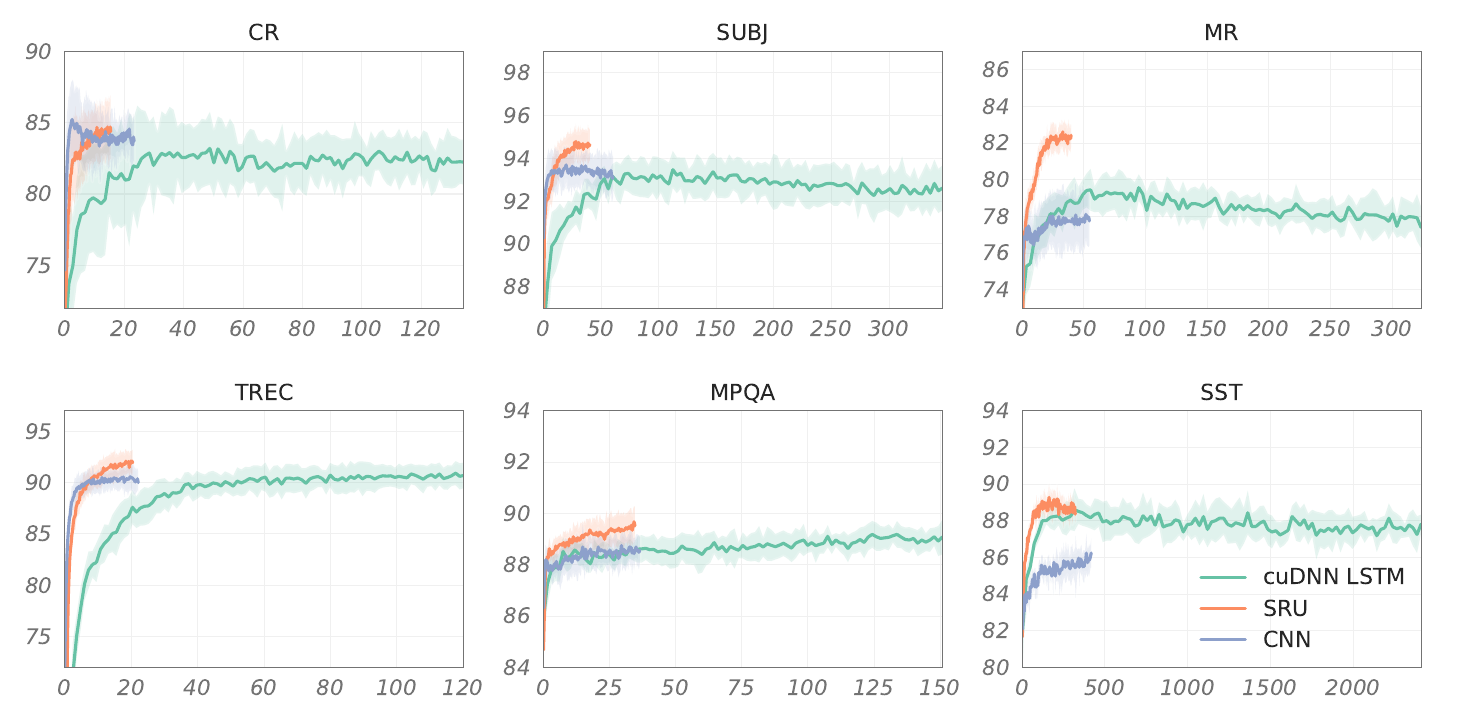}
\vspace{-0.05in}
\caption{\label{fig:clf}Mean validation accuracies (y-axis) and standard deviations of the CNN, 2-layer LSTM and 2-layer SRU models. We plot the curves of the first 100 epochs. X-axis is the training time used (in seconds). Timings are performed on NVIDIA GeForce GTX 1070 GPU, Intel Core i7-7700K Processor and cuDNN 7003.}
\end{figure*}

\paragraph{Setup} We stack multiple SRU layers and use the last output state to predict the class label for a given sentence.
We train for 100 epochs and use the validation (i.e., development) set to select the best training epoch.
We perform 10-fold cross validation for datasets that do not have a standard train-evaluation split.
The result on SST is averaged over five independent trials.
We use Adam~\citep{Kingma:14adam} with the default learning rate 0.001, a weight decay 0 and a hidden dimension of 128.

We compare SRU with a wide range of methods on these datasets, including various convolutional models~\citep{kalchbrenner:2014,Kim14,zhang2015sensitivity} and a hierarchical sentence model~\citep{zhao2015self} reported as the state of the art on these datasets~\citep{conneau17}.
Their setups are not exactly the same as ours, and may involve more tuning on word embeddings and other regularizations.
We use the setup of~\citet{Kim14} but do not fine-tune word embeddings and the learning method for simplicity.
In addition, we directly compare against three baselines trained using our code base: a re-implementation of the CNN model of~\citet{Kim14}, a two-layer LSTM model and Quasi-RNN~\citep{bradbury2016quasi}.
We use the official implementation of Quasi-RNN and also implement a version with highway connection for a fair comparison.
%In addition, we re-implement the CNN model of~\citet{Kim14} and a two-layer LSTM baseline to directly compare the model performance. 
These baselines are trained using the same hyper-parameter configuration as SRU.
%We also compare to the CNN model of~\citet{Kim14}, with the  same filter windows of $3$, $4$, and $5$ as the original work. 
%with the same number of filter width and number of filters as the original work. 
%We use the same filter widths and number of filters as~\citep{Kim14}.

\paragraph{Results} 
Table~\ref{table:clf} compares the test results on the six benchmarks. 
We select the best number reported in previous methods when multiple model variants were explored in their experiments.
Despite our simple setup, SRU outperforms most previous methods and achieves comparable results compared to the state-of-the-art but more sophisticated model of~\citet{zhao2015self}.
Figure~\ref{fig:clf} shows validation performance relative to training time for SRU, cuDNN LSTM and the CNN model.
Our SRU implementation runs 5--9 times faster than cuDNN LSTM, and 6--40\% faster than the CNN model of~\citet{Kim14}.
On the movie review (MR) dataset for instance, SRU completes 100 training epochs within 40 seconds, while LSTM takes over 320 seconds.
%For example, on the movie review task (MR), our model completes $100$ training epochs within $40$ seconds, while cuDNN LSTM takes more than $450$ seconds.

\subsection{Question Answering}
\label{sec:exp:qa}

\paragraph{Dataset}
We use the Stanford Question Answering Dataset~\citep[SQuAD;][]{rajpurkar2016squad}. 
SQuAD is a large machine comprehension dataset that includes over 100K question-answer pairs extracted from Wikipedia articles.
We use the standard train and development sets.%\footnote{\url{https://rajpurkar.github.io/SQuAD-explorer/}}.

\newcommand\Bstrut{\rule[-1.3ex]{0pt}{0pt}}   % = `bottom' strut
\begin{table*}[!ht!]
  \centering
  \begin{tabular}{@{~~}lccc@{~~~~~~}ccc@{~~}}
    \toprule
    \multirow{2}{*}{Model} & \multirow{2}{*}{\# layers} & \multirow{2}{*}{{~~~~}Size{~~~~}} & Dev & Dev & \multicolumn{2}{c}{Time per epoch{~}}\\
    & & & EM & F1 & {~}RNN{~} & {~}Total{~} \\
    \midrule
LSTM & \multirow{2}{*}{3} & \multirow{2}{*}{4.1m} & \multirow{2}{*}{69.5} & \multirow{2}{*}{78.8} & \multirow{2}{*}{316s} & \multirow{2}{*}{431s} \\
\citep{chen2017reading} & &  &  &  &  & \\
%Bi-LSTM & 4 & 5.8m & 69.6 & 78.9 & 729s & 872s \\
\midrule
\multirow{2}{*}{QRNN (k=1) + highway} & 4 & 2.4m & 70.1$\,\pm\,$0.1& 79.4$\,\pm\,$0.1& 113s & 214s \\
% & 5 & 2.8m & 70.7$\,\pm\,$0.2 & 80.0$\,\pm\,$0.2 \\
 & 6 & 3.2m & 70.6$\,\pm\,$0.1 & 79.6$\,\pm\,$0.2 & 161s & 262s \\
\midrule
SRU & 3 & 2.0m & \textbf{70.2}$\,\pm\,$0.3 & \textbf{79.3}$\,\pm\,$0.1 & 58s & 159s\\
SRU & 4 & 2.4m & \textbf{70.7}$\,\pm\,$0.1 & \textbf{79.7}$\,\pm\,$0.1 & 72s & 173s\\
SRU & 6 & 3.2m & \textbf{71.4}$\,\pm\,$0.1 & \textbf{80.2}$\,\pm\,$0.1 & 100s & 201s\\
    \bottomrule
   \end{tabular}
   %\vspace{0.05in}
   \caption{\label{table:squad}Exact match (EM) and F1 scores of various models on SQuAD (Section~\ref{sec:exp:qa}). 
   We also report the total processing time per epoch and the time spent in RNN computations.
   SRU outperforms other models, and is more than five times faster than cuDNN LSTM. }
\end{table*}

\paragraph{Setup}
We use the Document Reader model of \citet{chen2017reading} as our base architecture for this task.
The model is a combination of word-level bidirectional RNNs and attentions, providing a good testbed to compare our bidirectional SRU implementation with other RNN components.\footnote{The current state-of-the-art models~\citep{seo2016bidirectional,wang2017gated} make use of additional components such as character-level embeddings, which are not directly comparable to the setup of~\citet{chen2017reading}.
However, these models can potentially benefit from SRU since RNNs are incorporated in the model architecture.}

We use the open source implementation of Document Reader in our experiments.\footnote{\url{https://github.com/hitvoice/DrQA}}
We train models for up to 100 epochs, with a batch size of 32 and a hidden dimension of 128.
Following the author suggestions, we use the Adamax optimizer~\citep{Kingma:14adam} and variational dropout~\citep{Gal2016Theoretically} during training.
We compare with two alternative recurrent components:
the bidirectional LSTM adopted in the original implementation of \citet{chen2017reading} and Quasi-RNN with highway connections for improved performance.
%set filter width to 1 so it has the minimal (fastest) computation.
%We also add the highway connection to improve its performance for a fair comparison.
%We use a dropout of $0.5$ for input word embeddings, $0.2$ for SRU layers, and $0.3$ for LSTM layers.

\paragraph{Results}
Table~\ref{table:squad} summarizes the results on SQuAD.
SRU achieves 71.4\% exact match and 80.2\% F1 score, outperforming the bidirectional LSTM model by 1.9\% (EM) and 1.4\% (F1) respectively. 
SRU also exhibits over 5x speed-up over LSTM and 53--63\% reduction in total training time.
In comparison with QRNN, SRU obtains 0.8\% improvement on exact match and 0.6\% on F1 score, and runs 60\% faster.
\yae{This speed improvement highlights the impact of the fused kernel (Algorithm 1).
While the QRNN baseline involves a similar amount of computation, assembling all element-wise operations of both directions in SRU achieves better GPU utilization.}{}

\subsection{Machine Translation}
\label{sec:exp:mt}

\paragraph{Dataset}
We train translation models on the WMT English$\rightarrow$German dataset, a standard benchmark for translation systems~\citep{peitz-EtAl,li-EtAl:2014,jean15}.
The dataset consists of 4.5 million sentence pairs.
We obtain the pre-tokenized dataset from the OpenNMT project~\citep{opennmt17}.
The sentences were tokenized using the word-piece model~\citep{wu2016gnmt}, which generates a shared vocabulary of about 32,000 tokens.
Newstest-2014 and newstest-2017 are provided and used as the validation and test sets.\footnote{\url{https://github.com/OpenNMT/OpenNMT-tf/tree/master/scripts/wmt}}
%We use the WMT 2014 English$\rightarrow$German translation task.  
%We use the data and implementation available from the OpenNMT project~\citep{opennmt17}.
%We pre-process the training corpus following standard practice~\citep{peitz-EtAl,li-EtAl:2014,jean15}. 
%About 4M translation pairs are left after processing.
%The news-test-2014 data is used as the test set and the concatenation of news-test-2012 and news-test-2013  is used as the development set.
%The processed training and evaluation data are taken from~\citep{jean15}.

\paragraph{Setup}
We use the state-of-the-art Transformer model of~\citet{vaswani2017attention} as our base architecture.
In the base model, a single Transformer consists of a multi-head attention layer and a bottleneck feed-forward layer.
We substitute the feed-forward network using our SRU implementation:
\begin{align*}
\text{base:} &\quad \W\cdot\text{ReLU\_layer}(\x) + \bb\\
\text{ours:} &\quad \W\cdot\text{SRU\_layer}(\x) + \bb\;\;.
\end{align*}
The intuition is that SRU can better capture sequential information as a recurrent network, and potentially achieve better performance while requiring fewer layers.
%These changes are made on top of OpenNMT-py, the PyTorch version which 

We keep the model configuration the same as \citet{vaswani2017attention}:
the model dimension is $d_\text{model}=512$, the feed-forward and SRU layer has inner dimensionality $d_\text{ff}=d_\text{sru}=2048$, and positional encoding~\cite{gehring2017convolutional} is applied on the input word embeddings.
The base model without SRU has 6 layers, while we set the number of layers to 4 and 5 when SRU is added.
Following the original setup, we use a dropout probability 0.1 for all components, except the SRU in the 5-layer model, for which we use a dropout of 0.2 as we observe stronger over-fitting in training.

We use a single NVIDIA Tesla V100 GPU for each model.
The published results were obtained using 8 GPUs in parallel, which provide a large effective batch size during training.
To approximate the setup, we update the model parameters every 5$\times$5120 tokens and use 16,000 warm-up steps following OpenNMT suggestions.
We train each model for 40 epochs (250,000 steps), and perform 3 independent trials for each model configuration. A single run takes about 3.5 days with a Tesla V100 GPU.

\begin{table*}[!ht!]
\fontsize{10.5}{12.2}\selectfont
\centering
\begin{tabular}{lcccccc}
\toprule
\multirow{2}{*}{Model} & \multirow{2}{*}{\# layers} & \multirow{2}{*}{~~~Size~~~} & \multicolumn{2}{c}{BLEU score} & Speed & Hours \\
& & & Valid & Test & (toks/sec) & per epoch\\
\midrule
%OpenNMT-tf & \multirow{2}{*}{6} & \multirow{2}{*}{76m} & \multirow{2}{*}{26.9} & \multirow{2}{*}{28.0} & \multirow{2}{*}{-} & \multirow{2}{*}{-} \\[0.1em]
%(4 GPU training) \\[0.1em]
%\midrule
Transformer (base) & 6 & 76m & 26.6$\pm$0.2 (26.9) & 27.6$\pm$0.2 (27.9) & 20k & 2.0\\[0.4em]
% \midrule
Transformer (+SRU) & 4 & 79m & 26.7$\pm$0.1 (26.8) & 27.8$\pm$0.1 (28.3) & \textbf{22k} & \textbf{1.8}\\[0.3em]
Transformer (+SRU) & 5 & 90m & \textbf{27.1}$\pm$0.0 (\textbf{27.2}) & \textbf{28.3}$\pm$0.1 (\textbf{28.4}) & 19k & 2.1 \\[0.1em]
\bottomrule
\end{tabular}
\caption{English$\rightarrow$German translation results (Section~\ref{sec:exp:mt}).
We perform 3 independent runs for each configuration. 
We select the best epoch based on the valid BLEU score for each run, and report the average results and the standard deviation over 3 runs.
In addition, we experiment with averaging model checkpoints and use the averaged version for evaluation, following~\citep{vaswani2017attention}.
We show the best BLEU results achieved in brackets.
%As additional reference numbers, we include the results reported by OpenNMT Tensorflow implementation using 4 GPUs.
}
\label{table:nmt}
\end{table*}

\begin{figure}[!t!]
\centering
\includegraphics[width=2.9in]{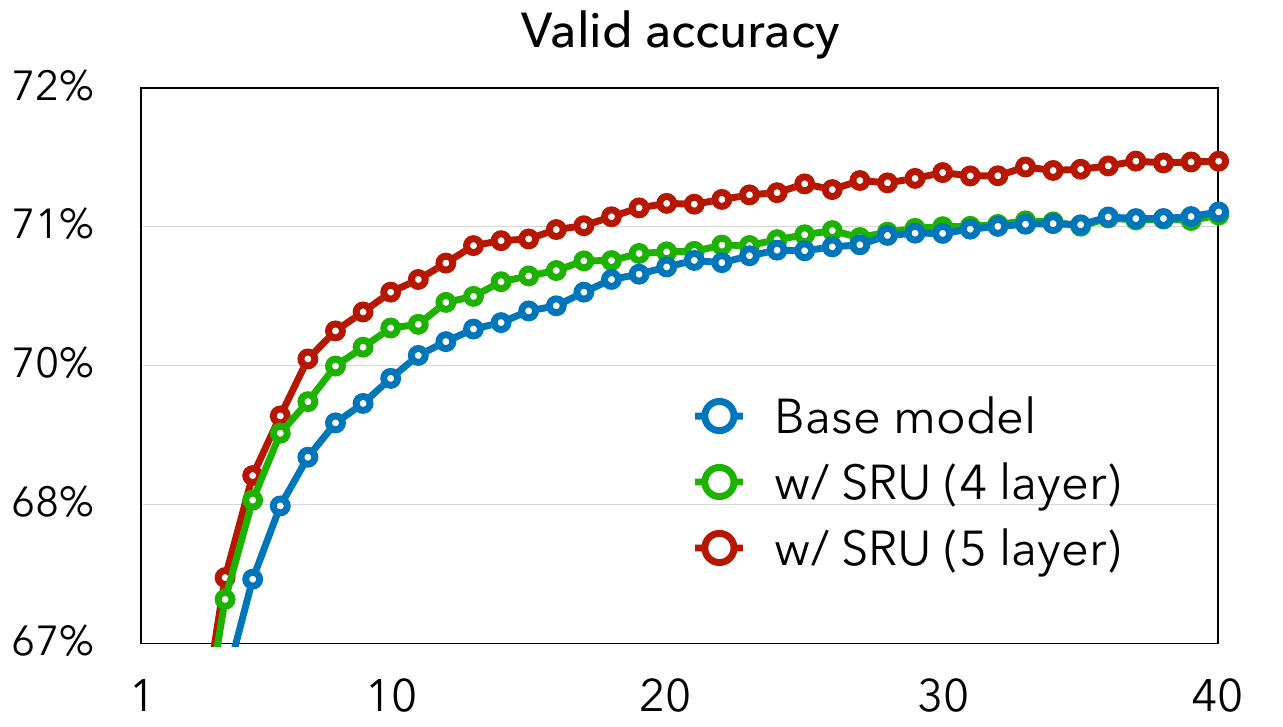}
\vspace{-0.05in}
\caption{Mean validation accuracy (y-axis) of different translation models after each training epoch (x-axis).}
\label{fig:mtacc}
\end{figure}

\paragraph{Results}
Table~\ref{table:nmt} shows the translation results. 
When SRU is incorporated into the architecture, both the 4-layer and 5-layer model outperform the Transformer base model.
For instance, our 5-layer model obtains an average improvement of 0.7 test BLEU score and an improvement of 0.5 BLEU score by comparing the best results of each model achieved across three runs. 
SRU also exhibits more stable performance, with smaller variance over 3 runs.
Figure~\ref{fig:mtacc} further compares the validation accuracy of different models. 
These results confirm that SRU is better at sequence modeling compared to the original feed-forward network (FFN), requiring fewer layers to achieve similar accuracy.
Finally, adding SRU does not affect the parallelization or speed of Transformer -- the 4-layer model exhibits 10\% speed improvement, while the 5-layer model is only 5\% slower compared to the base model.
We present more results and discussion in Appendix~\ref{sec:appendix:mt}.

\subsection{Character-level Language Modeling}
\label{sec:exp:lm}

\paragraph{Dataset}
We use Enwik8, a large dataset for character-level language modeling.
Following standard practice, we use the first 90M characters for training and the remaining 10M split evenly for validation and test.

\paragraph{Setup}
Similar to previous work, we use a batch size of 128 and an unroll size of 100 for truncated backpropagation during training.
We also experiment with an unroll size of 256 and a batch size of 64 such that each training instance has longer context.
We use a non-zero highway bias $\bb_r={-3}$ that is shown useful for training language model~\citep{ZillySKS17}. 
Previous methods employ different optimizers and learning rate schedulers for training. 
For simplicity and consistency, we use the Adam optimizer and the same learning rate scheduling (i.e., Noam scheduling) as the translation experiments.
We train a maximum of 100 epochs (about 700,000 steps).

We compare various recurrent models and use a parameter budget similar to previous methods.
In addition, we experiment with the factorization trick~\citep{kuchaiev2017factorization} to reduce the total number of parameters without decreasing the performance.
See details in Appendix~\ref{sec:appendix:experiments}.

\begin{table*}[!t!]
\fontsize{10.1}{12}\selectfont
\centering
\begin{tabular}{l@{~~~~~~}c@{~~~~~}c@{~~~~~}c@{~~~~~}cc@{~~~~~}c}
\toprule
Model & Size & \# layers & Unroll size & Valid & Test & Time \\
\midrule
\multicolumn{7}{@{~~}l}{\bf Best reported results:} \\[0.2em]
%LSTM~\cite{} & 21m & 7 & - & 1.67 & - \\
MI-LSTM~\cite{wu2016milstm} & 17m & 1 & 100 & - & 1.44 & - \\
HM-LSTM~\citep{ChungAB16} & 35m & 3 & 100 & - & 1.32 & - \\
%LSTM~\citep{melis2017state} & 27m & 4 & 1.29 & 1.31 & - \\
LSTM~\citep{melis2017state} & 46m & 4 & 50 & 1.28 & 1.30 & - \\
%RHN~\citep{ZillySKS17} & 21m & 10 & - & 1.30 & - \\
RHN~\citep{ZillySKS17} & 46m & 10 & 50 & - & 1.27 & - \\
FS-LSTM~\citep{mujika2017fast} & 47m & 4 & 100 & - & 1.25 & - \\
QRNN~\citep{merity2018analysis} & 26m & 4 & 200 & - & 1.33 & -\\
LSTM~\citep{merity2018analysis} & 47m & 3 & 200 & - & 1.23 & -\\%47min\\
\midrule
\multicolumn{7}{@{~~}l}{\bf Our setup:} \\[0.2em]
LSTM & 37m & 3 & 100 & 1.37 & 1.39 & 42min \\
LSTM & 37m & 6 & 100 & 1.35 & 1.38 & 48min \\
QRNN (k=1) & 37m & 6 & 100 & 1.36 & 1.38 & 30min \\[0.22em]
SRU & 37m & 6 & 100 & 1.29 & 1.30 & 28min \\
SRU & 37m & 10 & 100 & 1.26 & 1.27 & 29min \\
SRU (with projection) & 37m & 6 & 100 & 1.25 & 1.26 & 29min \\
SRU (with projection) & 47m & 8 & 100 & 1.21 & 1.21 & 39min \\
SRU (with projection) & 49m & 12 & 256 & 1.19 & 1.19 & 41min \\
\bottomrule
\end{tabular}
\caption{Validation and test BPCs of different recurrent models on Enwik8 dataset.
The last column presents the training time per epoch.
For SRU with projection, we set the projection dimension to 512.
\label{table:clm}
}
%We compare with a range of methods, including the previous state-of-the-art model of~\citep{ZillySKS17} and the LSTM model with hyper-parameter optimization~\citep{melis2017state}.}
\end{table*}

\paragraph{Results}
Table~\ref{table:clm} presents the results of SRU and other recurrent models. 
The 8-layer SRU model achieves validation and test bits per character (BPC) of 1.21, outperforming previous best reported results of LSTM, QRNN and recurrent highway networks (RHN).
Increasing the layer of SRU to 12 and using a longer context of 256 characters in training further improves the BPC to 1.19

\subsection{Ablation Analysis}
\label{sec:exp:ablation}

\begin{figure}[!t!]
\centering
\vspace{0.2in}
\includegraphics[width=3.1in]{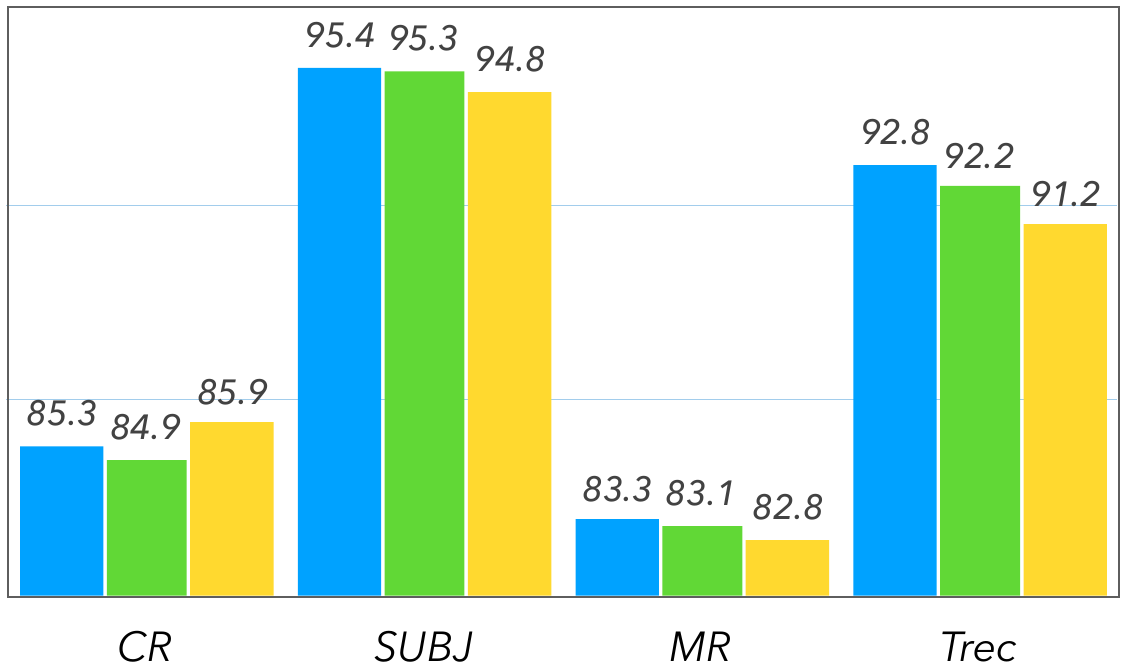}
\caption{Ablation analysis on the classification datasets.
Average validation results are presented. 
We compare the full SRU implementation (left blue), the variant without $\vv\odot\cc_{t-1}$ multiplication (middle green) and the variant without highway connection (right yellow).}
\label{fig:ablation}
\end{figure}

We perform ablation analyses on SRU by successively disabling different components:
\begin{itemize}
\item[(1)] Remove the point-wise multiplication term $\vv\odot\cc_{t-1}$ in the forget and reset gates. The resulting variant involves less recurrence and has less representational capacity.
\item[(2)] Disable the scaling correction by setting the constant $\alpha=1$.
\item[(3)] Remove the skip connections.% $(1-\rr_t)\odot \x_t$
\end{itemize}
We train model variants on the classification and question answering datasets.
Table~\ref{table:ablation} and Figure~\ref{fig:ablation} confirm the impact of our design decisions -- 
removing these components result in worse classification accuracies and exact match scores.

\begin{table}[!t!]
\centering
\begin{tabular}{lc@{~~~~~}c@{~~~}}
\toprule
Model & 4 layers & 6 layers \\
\midrule
SRU (full) &  70.7 & 71.4\\
  $\ \ -$ remove $\vv\odot\cc_{t-1}$ & 70.6 & 71.1\\
  $\ \ -$ remove $\alpha$-scaling & 70.3 & 71.0\\
  $\ \ -$ remove highway & 69.4 & 69.1\\
\bottomrule
\end{tabular}
%\begin{tabular}{lccc}
%\toprule
%\bf Model & SUBJ & MR & Trec \\
%\midrule
%SRU (full) &  95.4 & 83.3 & 92.8\\
%  $\ \ -$ remove $\vv\odot\cc_{t-1}$ & 95.3 & 83.1 & 92.2\\
%  $\ \ -$ remove $\alpha$-scaling & 94.8 & 82.2 & 92.3\\
%\bottomrule
%\end{tabular}
\caption{Ablation analysis on SQuAD. Components are successively removed and the EM scores are averaged over 4 runs.}
\label{table:ablation}
\end{table}

\section{Discussion}
This work presents Simple Recurrent Unit (SRU), a scalable recurrent architecture that operates as fast as feed-forward and convolutional units.
We confirm the effectiveness of SRU on multiple natural language tasks ranging from classification to translation.
We open source our implementation to facilitate future NLP and deep learning research.

\paragraph{Trading capacity with layers}
SRU achieves high parallelization by simplifying the hidden-to-hidden dependency.
This simplification is likely to reduce the representational power of a single layer and hence should be balanced to avoid performance loss. 
However, unlike previous work that suggests additional computation (e.g., n-gram filters) within the layer~\citep{BalduzziG16,bradbury2016quasi}, we argue that increasing the depth of the model suffices to retain modeling capacity.
Our empirical results on various tasks confirm this hypothesis.

%\paragraph{When will SRU be significantly faster.}
%The parallelization benefit of SRU becomes significant when the amount of computation per layer is relatively small to the GPU capacity.
%This can be observed in our experiments (1) when the recurrent model is small, as in the classification and question answering tasks, and (2) when stacking a lot layers given the same computation budget (so each layer involves less computation), as in the translation and language modeling tasks.

\section*{Acknowledgement}
%\ya{if submission is anonymized, must remove ack section}
We thank Alexander Rush and Yoon Kim for help with machine translation experiments, and Danqi Chen for help with SQuAD experiments.
We thank Adam Yala, Howard Chen, Jeremy Wohlwend, Lili Yu, Kyle Swanson and Kevin Yang for providing useful feedback on the paper and the SRU implementation.
A special thanks to Hugh Perkins for his support on the experimental environment setup and Runqi Yang for answering questions about his code.%, and the PyTorch community for enabling flexible neural module implementation.

\bibliography{emnlp2018}
\bibliographystyle{acl_natbib_nourl}

\clearpage
\newpage
\appendix
\section{Parameter Initialization Derivation}
\label{sec:appendix:init}

Following the derivation of Glorot and Kaiming initialization~\cite{glorot2010understanding,he2015delving}, we assume the values of each input vector $\x_t$ are i.i.d with zero mean and a small variance:
\begin{align*}
\text{E}[x_{t,i}] = 0, \quad \text{Var}[x_{t,i}] \ll 1 & & \forall 1\leq i\leq d\;\;.
\end{align*}
%Note that two vectors in the input sequence, say $\x_t$ and $\x_t'$, are not necessarily independent (because two words in an input sentence can be correlated).
%The correlation between input vectors will impact the scale of $\cc_t$ as we will show later.
We initialize each weight matrix with zero mean and a variance of $1/d$.
After a matrix multiplication $\mathbf{y} = \W\x_t$, each value $y_i$ would have
\begin{align*}
& \text{E}[y_i] = \text{E}[\sum_j w_{i,j}\x_{t,j}] = 0 \\
& \text{Var}[y_i] = \sum_j \text{Var}[w_{i,j}]\cdot \text{Var}[x_{t,j}] = \text{Var}[x] \;\;,
\end{align*}
which means the scale of the values after matrix multiplication remains the same.

\subsection{Computing Var[$\cc_t$]}
% Now consider a dimension of the state, say $i$, 
Let $f_{t,i}$ be the $i$-th entry of the forget gate $\f_t$:
\begin{align*}
f_{t,i} = \sigma(\w_{f,i}^\top \x_t + v_{f,i}\,c_{t-1,i} + b_{f,i})\;\;.
\end{align*}
The pre-activation value will be sufficiently close to 0 because the parameters are initialized with zero mean and small variance and the bias value is initially 0.
As a result,
\begin{align*}
E[f_{t,i}] = \sigma(0) = 0.5\;\;.
\end{align*}
The state value $c_{t,i}$ is computed according to
\begin{align*}
c_{t,i} = f_{t,i}\cdot c_{t-1,i} + (1-f_{t,i})\cdot (\w_i^\top \x_t)\;\;.
\end{align*}
Substituting the expectation of $f_{t,i}$ in, we get:\footnote{We are ignoring the correlation between $\f_{t,i}$ and $\f_{t,i'}$ here because their variance is small.}
\begin{align*}
c_{t,i} = \w_i^\top \left(\frac{\x_t}{2} + \frac{\x_{t-1}}{4} + \frac{\x_{t-2}}{8} + \cdots \right)\;\;.
\end{align*}
Therefore, $\text{E}[c_{t,i}]=0$ as $\text{E}[\w^\top\x]=0$.
The variance of $c_{t,i}$ however depends on the correlation between input vectors.
When the input vectors are independent: 
\begin{align*}
\text{Var}[c_{t,i}] &= \text{Var}[\w_i^\top \x]\left(\frac{1}{2^2} + \frac{1}{4^2} + \frac{1}{8^2} + \cdots\right)\\
& \approx \text{Var}[\w_i^\top \x] \cdot \frac{1}{3} = \text{Var}[x] / 3\;\;.
\end{align*}
However, the two vectors in the input sequence, for instance $\x_t$ and $\x_t'$, are not necessarily independent, for example because two words in an input sentence are often correlated.
When the input vectors are perfectly correlated $\x_t=\x_{t-1}=\cdots=\x$, on the other hand, 
\begin{align*}
\text{Var}[c_{t,i}] = \text{Var}[\w_i^\top \x] = \text{Var}[x]\;\;.
\end{align*}

\begin{figure}[!t!]
\vspace{0.1in}
%\centering
\includegraphics[width=3.1in]{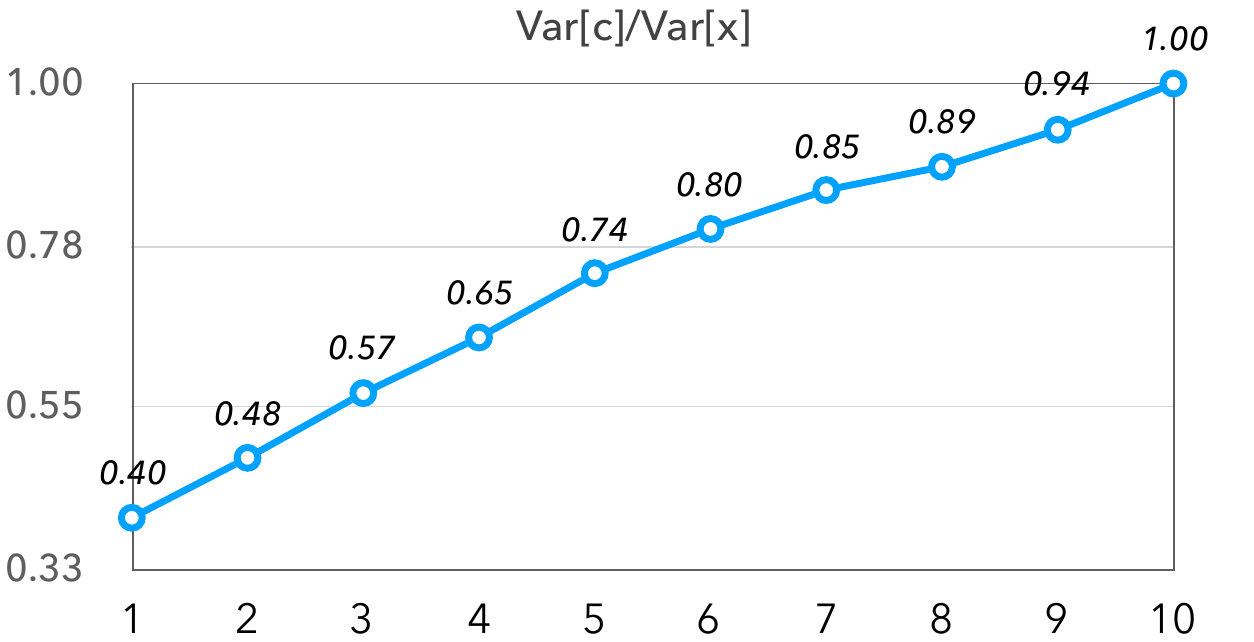}
\caption{Empirical estimation of the variance ratio $\text{Var}[\cc_t]/\text{Var}[\x_t]$ at each layer in a randomly initialized SRU model.
We use the pre-trained word2vec embeddings as input, resulting an initial ratio slightly higher than $1/3$.
As expected, the ratio increases to 1 in deep layers.
}
\label{fig:var}
\end{figure}

In practice, multiple SRU layers are stacked to construct a deep network.
The internal state $\cc_t$ and $\h_t$ would be a weighted combination of inputs $\{\x_1\cdots \x_t\}$, which will increase the correlation of the state vectors at different steps.
These state vectors are again fed into the next layer, and keep increasing the correlation. 
As a result, we expect the actual ratio between the variance of $\cc_t$ and that of the input of the current layer $\x_t$ lies between the two derived values,
\begin{align}
\frac{1}{3} \ \leq\ \frac{\text{Var}[c]}{\text{Var}[x]}\ \leq\ 1\;\;,
\end{align}
and would finally converge to the upper bound value of 1.
Figure~\ref{fig:var} confirms our expectation by computing the empirical value of $\text{Var}[c]/\text{Var}[x]$ in deep SRU networks.

\subsection{Computing Var[$\h_t$]}
Given the result in Equation (5), we  proceed to compute $\text{Var}[\h_t]$.
The $i$-th entry of $\h_t$ is similarly computed as
\begin{align*}
& h_{t,i} = r_{t,i}\cdot c_{t,i} + (1-r_{t,i})\cdot x_{t,i} \\
\text{where}\ \ \ \ & r_{t,i} = \sigma(\w_{r,i}^\top \x_t + v_{r,i} c_{t-1,i} + b_{r,i})\;\;.
\end{align*}
The highway reset gate is not necessarily initialized with a zero bias.
Let the initial bias be $b$ and $u=\w_{r,i}^\top \x_t + v_{r,i} c_{t-1,i}$ denote the rest of terms in the sigmoid function.
We have $\text{E}[u]=0$ and $\text{Var}[u] \ll 1$ because $\x_t$ and $\cc_{t-1}$ have small variance.

We approximate the value of $r_{t,i}$ using its Taylor expansion at $u=0$:
\begin{align*}
r_{t,i}&\ = \ \sigma(u+b) \\
&\ \approx\ \frac{e^b}{e^b+1} + \frac{e^b \cdot u}{(e^b+1)^2} \\
\text{E}[r_{t,i}^2]&\ \approx\ \frac{e^{2b}}{(e^b+1)^2} + \frac{e^{2b}\cdot u^2}{(e^b+1)^4}\;\;.
\end{align*}
We can ignore the term with $u^2$ since $\text{Var}[u]\ll 1$, which gives us
\begin{align*}
\text{E}[r_{t,i}^2]&\ \approx\ \frac{e^{2b}}{(e^b+1)^2}\;\;.
\end{align*}
Substituting this result in $\text{Var}[h_{t,i}]$,
\begin{align}
\text{Var}[h_{t,i}] &\ =\ \text{E}\left[r_{t,i}^2\cc_{t,i}^2 + (1-r_{t,i})^2 x_{t,i}^2 \right] \nonumber \\[0.5em]
&\ =\ \frac{e^{2b}\cdot \text{Var}[c]}{(e^b+1)^2} + \frac{\text{Var}[x]}{(e^b+1)^2}
\end{align}
Since from (5) we have $\text{Var}[x]/3 \leq \text{Var}[c] \leq \text{Var}[x]$, we get the bound of $\text{Var}[h_{t,i}]$
\begin{align*}
\frac{e^{2b}+3}{3(e^b+1)^2}\ \leq\ \frac{\text{Var}[h]}{\text{Var}[x]}\ \leq\ \frac{e^{2b}+1}{(e^b+1)^2}
\end{align*}
which is equivalent to 
\begin{align*}
\frac{1}{3}\ \leq\ \frac{\text{Var}[h]}{\text{Var}[x]}\ \leq\ \frac{1}{2}
\end{align*}
when $b=0$.

\subsection{Computing the Scaling Constant $\alpha$}
Finally, we compute the scaling constant $\alpha$ (Section~\ref{sec:method:init}).
Using the result in Equation (6), when $\alpha$ is introduced we get:
\begin{align*}
\text{Var}[h_{t,i}] &\ =\ \frac{e^{2b}\cdot \text{Var}[c]}{(e^b+1)^2} + \frac{\alpha^2\cdot \text{Var}[x]}{(e^b+1)^2} \\[0.5em]
&\ \approx\ \frac{e^{2b} + \alpha^2}{(e^b+1)^2} \cdot \text{Var}[x]\;\;,
\end{align*}
as $\text{Var}[c]\rightarrow \text{Var}[x]$ according to Equation (5) and the empirical evaluation (Figure~\ref{fig:var}).
This implies $e^{2b} + \alpha = (1+e^b)^2$ if we want $\text{Var}[h]\approx \text{Var}[x]$.
By solving for $\alpha$ we have
\begin{align*}
\alpha\ = \ \sqrt{1+2\cdot e^b}\;\;,
\end{align*}
and $\alpha=\sqrt{3}$ when $b=0$.

\section{Experimental Details}
\label{sec:appendix:experiments}
We include additional experimental setup and results in this section.

\subsection{Classification}
The data and pre-processing code are obtained from the code repository of Harvard NLP.\footnote{\url{https://github.com/harvardnlp/sent-conv-torch}}

We use a batch size of 32 and a dropout probability of 0.5 for all models.
In addition, we increment the dropout to 0.55 or 0.6 for the 8-layer SRU model.
Following the implementation of~\citep{Kim14}, out-of-vocabulary words that are not in the pre-trained embeddings are initialized with random vectors with values from $[-0.25, 0.25]$.

\subsection{Question Answering}
We use a word embedding dropout of 0.5 and a recurrent dropout of 0.2.
In the setup of~\citet{chen2017reading}, the bi-LSTM models concatenates the output of each layer and feed it to subsequent layers.
This helps the gradient propagation and improves the final performance.
With highway connection, this is no longer necessary.
In SRU and Q-RNN (with highway), only the output of the last layer is given to subsequent layers.

\subsection{Machine Translation}
\label{sec:appendix:mt}
We use the OpenNMT PyTorch implementation for the translation experiments.%\footnote{\url{https://github.com/OpenNMT/OpenNMT-py}}.
Table~\ref{table:option} shows the list of configuration options used for training.
For evaluation, we use beam size 5 and length penalty 0.6. 

\begin{table}[!h!]
\centering
\begin{tabular}{|@{~~}ll|l|}
% \toprule[0.1em]
\hline
-layers & 4 to 6 & -share\_embedding \\
-rnn\_size & 512 & -position\_encoding \\
-word\_vec\_size & 512 & -param\_init 0\\
-batch\_type & tokens & -max\_grad\_norm 0 \\
-normalization & tokens & -dropout 0.1\\
-batch\_size & 5120  & -label\_smoothing 0.1\\
-accum\_count & 5 & -epoch 40\\
-optim & adam & -param\_init\_glorot \\
-learning\_rate & 2 & \\
-adam\_beta2 & 0.998 & \\
-decay\_method & noam & \\
-warmup\_steps & 16000 & \\
\hline
% \bottomrule[0.1em]
\end{tabular}
\caption{Translation training configuration.}
\label{table:option}
\end{table}
\begin{table*}[!t!h!]
\fontsize{10.5}{12.2}\selectfont
\centering
\begin{tabular}{@{~~~}c@{~~~}|rc|rc|rc}
\toprule
\multirow{2}{*}{Epoch} & \multicolumn{2}{c|}{{~~~~}Transformer base{~~~~}} & \multicolumn{2}{c|}{{~~~~}w/ SRU (4 layer){~~~~}} & \multicolumn{2}{c}{{~~~~}w/ SRU (5 layer){~~~~}} \\
& {~~}Valid & Test & {~~}Valid & Test & {~~}Valid & Test \\
\midrule
20 & 26.1 & 27.3 & 26.2 & 27.6 & 26.6 & 27.9\\
21 & 26.2 & 27.3 & 26.3 & 27.7 & 26.6 & 28.1\\
22 & 26.1 & 27.4 & 26.3 & 27.8 & 26.7 & 28.0\\
23 & 26.2 & 27.4 & 26.4 & 27.7 & 26.8 & 28.1\\
24 & 26.2 & 27.4 & 26.4 & 27.8 & 26.7 & 28.0\\
25 & 26.3 & 27.4 & 26.4 & 27.7 & 26.6 & 28.1\\
26 & 26.5 & 27.5 & 26.5 & 27.7 & 26.7 & 28.1\\
27 & 26.4 & 27.6 & 26.4 & 27.6 & 26.8 & 28.1\\
28 & 26.4 & 27.6 & 26.4 & 27.7 & 26.7 & 28.2\\
29 & 26.4 & 27.5 & 26.4 & 27.8 & 26.8 & 28.2\\
30 & 26.5 & 27.7 & 26.4 & 27.8 & 26.9 & 28.1\\
31 & 26.4 & 27.6 & 26.6 & 27.7 & 26.9 & 28.3\\
32 & 26.5 & 27.5 & 26.5 & 27.8 & 26.9 & 28.3\\
33 & 26.5 & 27.5 & 26.5 & 27.8 & 27.1 & 28.3\\
34 & 26.4 & 27.6 & 26.5 & 27.9 & 26.9 & 28.2\\
35 & 26.4 & 27.6 & 26.5 & 27.9 & 26.9 & 28.2\\
36 & 26.5 & 27.6 & 26.5 & 27.8 & 26.9 & 28.3\\
37 & 26.5 & 27.5 & 26.5 & 27.8 & 26.9 & 28.2\\
38 & 26.5 & 27.6 & 26.5 & 28.0 & 27.0 & 28.2\\
39 & 26.5 & 27.6 & 26.7 & 27.8 & 27.0 & 28.2\\
40 & 26.6 & 27.6 & 26.6 & 27.9 & 27.0 & 28.2\\
\bottomrule
\end{tabular}
\caption{Average BLEU scores after each epoch.}
\label{table:full_mt}
\end{table*}
\begin{figure}[ht]
%\centering
\includegraphics[width=3in]{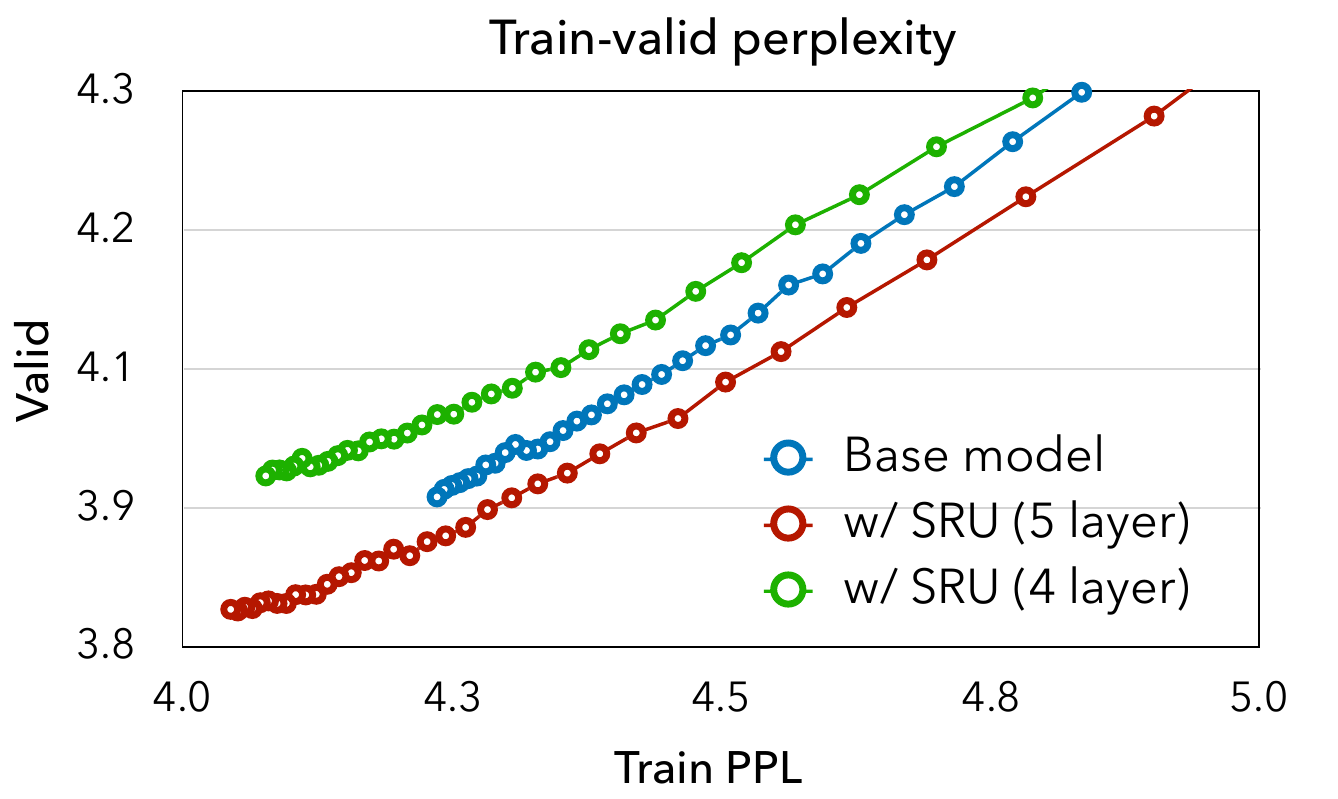}
\caption{Training and validation perplexity curves of the base model and two SRU models.}
\label{fig:mtppl}
\end{figure}

Table~\ref{table:full_mt} shows the averaged BLEU score of each model from 20th to 40th epoch.
The improvement over the Transformer base model is consistent across different epochs.

Figure~\ref{fig:mtppl} plots the training and validation perplexity of three models. 
With a higher dropout (0.2) used for the SRU, the 5-layer model gets consistent lower validation perplexity over the base model and the 4-layer model.
We also see that models with SRU exhibit much faster training progress with much lower training perplexity, suggesting the models could be tuned better with further training regularization.

\subsection{Character-level Language Modeling}
We train all models using a weight decay of $10^{-7}$ and a gradient clipping of $0.3$.
We set the learning rate factor of Noam scheduling to $3$ and the warmup steps to $32,000$.
We tune the dropout probability from $\{0.2, 0.3\}$.

The projection (bottleneck) trick is implemented as follows.
Recall that the batched multiplication of SRU is computed as
\begin{align}
\nonumber \left(\begin{array}{l}\W \\ \W_f \\ \W_r \end{array}\right) [\x_1, \x_2, \cdots, \x_L] \;\;.
\end{align}
The stacked parameter matrices on the left is re-parameterized by a low-rank factorization,
\begin{align}
\nonumber \left(\begin{array}{l}\W \\ \W_f \\ \W_r \end{array}\right) = \mathbf{P}^\top\mathbf{Q} \;\;,
\end{align}
where $\mathbf{Q} \in \mathbb{R}^{d_{in}\times d'}$ and $\mathbf{P} \in \mathbb{R}^{3d_{out} \times d'}$ are two new parameter matrices to be learned, and $d'$ is the projection dimension that is much smaller than the input and output dimension of the SRU.

\end{document}